\documentclass{article}

\usepackage[preprint]{neurips_2026}

\workshoptitle{AI Native Academia}

\usepackage[utf8]{inputenc} 
\usepackage[T1]{fontenc}    
\usepackage{hyperref}       
\usepackage{url}            
\usepackage{booktabs}       
\usepackage{amsfonts}       
\usepackage{nicefrac}       
\usepackage{microtype}      
\usepackage[normalem]{ulem}
\usepackage{xcolor}         
\usepackage{natbib}
\usepackage{pifont}
\usepackage{adjustbox}
\newcommand{\datasetname}{Metag}
\usepackage{multicol, multirow}
\usepackage{listings}
\usepackage{graphicx}
\usepackage{tabularx}
\usepackage[table]{xcolor}
\usepackage{arydshln}
\usepackage[most]{tcolorbox}
\definecolor{codegreen}{rgb}{0,0.6,0}
\definecolor{codegray}{rgb}{0.5,0.5,0.5}
\definecolor{codepurple}{rgb}{0.58,0,0.82}
\definecolor{backcolour}{rgb}{0.95,0.95,0.92}
\makeatletter
\newcommand{\datasetavailability}{%
  \@ifpackagewith{neurips_2026}{preprint}
    {The dataset is publicly available at
     \url{https://github.com/microsoft/Metag-dataset}}
    {The dataset will be made publicly available upon acceptance}%
}
\makeatother

\lstdefinestyle{mystyle}{
    backgroundcolor=\color{backcolour},   
    commentstyle=\color{codegreen},
    keywordstyle=\color{magenta},
    numberstyle=\tiny\color{codegray},
    stringstyle=\color{codepurple},
    basicstyle=\ttfamily\footnotesize,
    breakatwhitespace=false,         
    breaklines=true,                 
    captionpos=b,                    
    keepspaces=true,                 
    numbers=left,                    
    numbersep=5pt,                  
    showspaces=false,                
    showstringspaces=false,
    showtabs=false,                  
    tabsize=2
}

\lstdefinelanguage{json}{
    showstringspaces=false,
    morestring=[b]",
    stringstyle=\color{codegreen},
    commentstyle=\color{codepurple},
    literate=
     *{0}{{{\color{magenta}0}}}{1}
      {1}{{{\color{magenta}1}}}{1}
      {2}{{{\color{magenta}2}}}{1}
      {3}{{{\color{magenta}3}}}{1}
      {4}{{{\color{magenta}4}}}{1}
      {5}{{{\color{magenta}5}}}{1}
      {6}{{{\color{magenta}6}}}{1}
      {7}{{{\color{magenta}7}}}{1}
      {8}{{{\color{magenta}8}}}{1}
      {9}{{{\color{magenta}9}}}{1}
      {:}{{{\color{black}{:}}}}{1}
      {,}{{{\color{black}{,}}}}{1}
      {\{}{{{\color{black}{\{}}}}{1}
      {\}}{{{\color{black}{\}}}}}{1}
      {[}{{{\color{black}{[}}}}{1}
      {]}{{{\color{black}{]}}}}{1},
      }

\title{\datasetname{}: A dataset to build agentic meta-reviewing capabilities}

\author{%
  \textbf{Anirudh Sundar}
  \textsuperscript{1}\thanks{Main Contribution}
  \quad \textbf{Min Chen}\textsuperscript{1}
  \quad \textbf{Divya Tadimeti}\textsuperscript{1}
  \quad \textbf{Gemma Zhang}\textsuperscript{1}
  \quad \textbf{Xinyi Alice Li}\textsuperscript{1}
  \\
  \textbf{Nigel Boachie Kumankumah}\textsuperscript{1}
  \quad \textbf{Pavan Uttej Ravva}\textsuperscript{1}
  \quad \textbf{Sadid Hasan}\textsuperscript{1}
  \quad \textbf{Somya Chatterjee}\textsuperscript{1} \\
  \textbf{Pruthvi Prakash Navada}\textsuperscript{1}
  \quad \textbf{Xiao Wang}\textsuperscript{1}
  \quad \textbf{Yue Kang}\textsuperscript{1}
  \quad \textbf{Sulaiman Vesal}\textsuperscript{1}
  \quad \textbf{Larry Heck}\textsuperscript{2}
  \\[0.5em]
  \textsuperscript{1}Microsoft
  \qquad
  \textsuperscript{2}Georgia Institute of Technology
  \\
  \texttt{\{anisundar, svesal\}@microsoft.com, larryheck@gatech.edu}
}

\begin{document}

\maketitle

\begin{abstract}
AI tools increasingly support tasks across the scientific research cycle, from experiment design and manuscript preparation to peer review. At the same time, the continuing growth in conference submissions has increased the burden on meta-reviewers, who must keep up with reviewer feedback, author rebuttals, and manuscript revisions. To address this concern, this paper introduces \datasetname{}, a human-annotated dataset to accelerate the development of meta-reviewing agents, specifically to identify changes made to scientific articles during the review-rebuttal process. 
Each instance contains a reviewer concern, the author's proposed resolution, and the manuscript diffs implementing the stated change.
\datasetname{} is collected by obtaining manuscript versions from before the review deadline and after acceptance, computing differences between the two documents, and asking human annotators to align these differences with action items from OpenReview discussions.  The resulting dataset consists of 349 high-quality action items tied to paper differences and will enable building methods to empower meta reviewers to quickly identify whether authors have addressed reviewer statements and where in the paper those changes have been made, resulting in additional transparency and traceability throughout peer review. \datasetavailability{}. 
\end{abstract}

\section{Introduction}
\label{sec:introduction}

Recent work has seen an increase in the usage of AI tools and assistants both in authoring and reviewing scientific papers \cite{Liang2024MonitoringAC}. For authors, AI assistants help in proofreading papers, improving writing structure, and rectifying grammatical errors. Similarly, AI tools have also been developed to assist in the reviewing process. Paper Assistant Tool \cite{icml2026pat}, an AI tool introduced in the ICML 2026 review process, has been deployed to assist in flagging issues, highlighting errors in methods, and helping improve the writing of papers. AI tools reduce cognitive burden across experimentation and reviewing, thereby accelerating the research cycle and resulting in massive growth in the scale of submissions to premier AI conferences \cite{bok2025openreview}. While the development of such tools have primarily targeted paper authors and reviewers, an important yet relatively under-explored area of research is the development of AI tools to assist the meta-reviewing process \cite{kuznetsov2024can}.  

Meta-reviewers must synthesize large volumes of information across reviews, rebuttals, and author discussions into a coherent statement. This process is challenging because the relevant evidence is distributed across long, interconnected texts, reviewers may emphasize different concerns, and author responses may describe revisions without making it immediately clear where those revisions appear in the final manuscript \cite{kuznetsov2024can}. Although NLP systems can assist with review aggregation and summarization, effective support for meta-reviewers requires traceability, the ability to connect concerns raised during peer review to the concrete changes made by authors.

\begin{figure}[h]
    \centering
    \includegraphics[width=\linewidth]{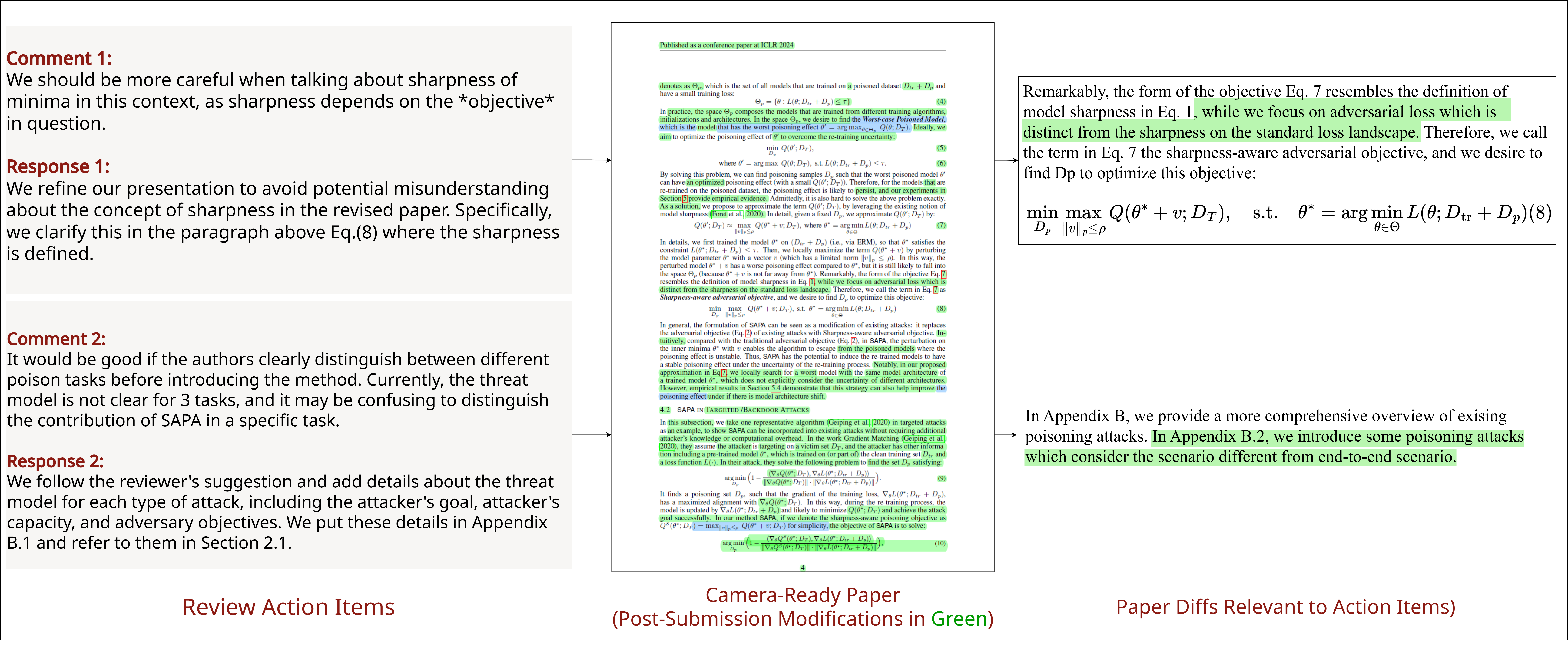}
    \caption{An example of the data collection process and sample data from \datasetname{}. Based on the action items identified by reviewers and the modifications made post submission to create the Camera-Ready paper (edits highlighted in green), the task is to extract the relevant diffs specific to the action items.}
    \label{fig:teaser}
\end{figure}

Such traceability can improve transparency and accountability in the review-rebuttal process. It can help meta-reviewers and program committee members verify whether promised revisions were implemented, distinguish substantive changes from unrelated edits, and inspect the manuscript evidence underlying an author's response. Existing tools can identify textual differences between versions of a PDF \cite{ssibb_pdf_diff_viewer}, but they do not generally explain which reviewer action item motivated each change. Automatically linking manuscript diffs to reviewer concerns therefore represents an important and underexplored extension of document comparison.

To support research on this problem, this paper introduces a data-collection methodology and releases \datasetname{}, a dataset that links action items extracted from reviewer-author dialogues to their corresponding changes between original and revised scientific papers. \datasetname{} is intended to support the development of agentic human-in-the-loop systems that assist meta-reviewers navigate revisions by verifying author statements and ground judgments in manuscript evidence. An example of the dataset is provided in Figure~\ref{fig:teaser}, given the action items from the review on OpenReview, the task is to pick out the specific changes that effectuate the reviewer request.


\datasetname{} is built by computing a diff between an accepted paper's pre-review arXiv submission against its camera-ready version, then linking specific edits to the reviewer comments that prompted them via reviewer-author dialogue. The process yields a dataset of paired edits and comments connecting reviewer feedback to the concrete manuscript changes addressing them. \datasetavailability{}  under the CDLA-2.0 license \footnote{\url{https://cdla.dev/permissive-2-0/}}. 

\section{Related Work}

\paragraph{Scientific reviews without document edits.}
Several datasets study scientific peer review without modeling how authors revise their manuscripts. \textsc{cPAPERS} \cite{sundar2024cpapers} and \textsc{SciDQA} \cite{singh-etal-2024-scidqa} support question answering grounded in scientific papers and their OpenReview discussions, while \textsc{ORSUM} \cite{zeng2024scientific} focuses on scientific opinion summarization. \textsc{Revise and Resubmit} \cite{kuznetsov-etal-2022-revise} introduces pragmatic tagging of review sentences and links reviewer suggestions to manuscript sections through explicit section references and embedding similarity. These resources capture the content, intent, or organization of reviews, but do not align reviewer requests with changes observed between manuscript versions. Similarly, \textsc{Friction} \cite{10.1145/3706598.3714316} uses LLMs to identify feedback for novice writers and presents a heatmap indicating passages that may require revision. Although it supports feedback-driven writing, it does not model revisions arising from scientific peer review.

\paragraph{Document edits without scientific reviews.}
A complementary line of work studies iterative document revision independently of peer review. \textsc{arXivEdits} \cite{jiang2022arxivedits} provides a computational framework for extracting changes across arXiv versions at the document, sentence, and word levels. They additionally classify edit intentions, enabling analysis of why revisions were made. \textsc{IteraTeR}~\cite{du-etal-2022-understanding-iterative} collects edits from arXiv, Wikipedia, and Wikinews and introduces edit-intention classification, but does not include the reviewer feedback that prompted scientific revisions. \textsc{TETRA} \cite{mita2024towards} also addresses document revision, using edits produced by professional editors rather than changes made in response to scientific peer review. These datasets model what changed, but generally cannot connect those changes to reviewer concerns or author commitments.

\paragraph{Scientific reviews and document edits.}
The work most closely related to \datasetname{} is \textsc{ARIES} \cite{darcy-etal-2024-aries}, which aligns reviewer comments with revisions to scientific manuscripts. ARIES relies on annotators to identify actionable review comments and align them directly with relevant textual spans. In contrast, \datasetname{} first computes structured differences between the original and revised PDFs and then asks annotators to select the diffs that implement each action item extracted from the reviewer-author dialogue. This formulation preserves insertions, deletions, replacements, page locations, and surrounding context, while framing the task as selecting relevant edits from the complete set of manuscript changes. 

More broadly, \cite{kuznetsov2024can} provides a comprehensive taxonomy of NLP research across the peer-review pipeline, including methods that jointly analyze manuscripts, reviews, and reviewer-author discussions. \datasetname{} complements this literature by focusing specifically on traceability between review dialogue and observable manuscript revision. A comparison of the different related datasets in this domain is provided in Table \ref{tab:dataset-comaprison}. 

\begin{table}[t]
    \centering
    \caption{Comparison between different datasets targeting iterative edits and scientific reviews. \#~Samples = number of samples verified by human annotators.}
    \begin{tabular}{l r c c c}
    \toprule
    Dataset & \# Samples & Scientific Reviews & Edits & Task  \\
    \midrule 
    cPAPERS \cite{sundar2024cpapers} & 5,030 & \checkmark & \ding{55} & Question-Answering \\
    SciDQA \cite{singh-etal-2024-scidqa} & 2,937 & \checkmark & \ding{55} & Question-Answering \\
    \textsc{ORSUM} \cite{zeng2024scientific} & 57,536 & \checkmark & \ding{55} & Opinion Summarization \\ 
    Revise and Resubmit   \cite{kuznetsov-etal-2022-revise}  &  21,289 & \checkmark & \ding{55} & Edit Intent Class.\\
    \textsc{TETRA} \cite{mita2024towards} &  1,368 & \ding{55} & \checkmark & Document Revision \\
    \textsc{arXivEdits} \cite{jiang2022arxivedits} &13,008 & \ding{55} & \checkmark & Edit Intent Class.  \\
    \textsc{IteraTeR}  \cite{du-etal-2022-understanding-iterative} & 559 & \ding{55} & \checkmark & Document Revision \\
    \textsc{ARIES} \cite{darcy-etal-2024-aries} & 131 & \checkmark & \checkmark & Document Revision\\
    \midrule
    \datasetname{} (Ours) & 349 & \checkmark & \checkmark & Document Revision\\
    \bottomrule
    \end{tabular}
    \label{tab:dataset-comaprison}
\end{table}

\section{Method}

\subsection{Dataset Collection}
 \label{sec:data-collection}

\begin{figure}[h]
    \centering
    \includegraphics[width=\linewidth]{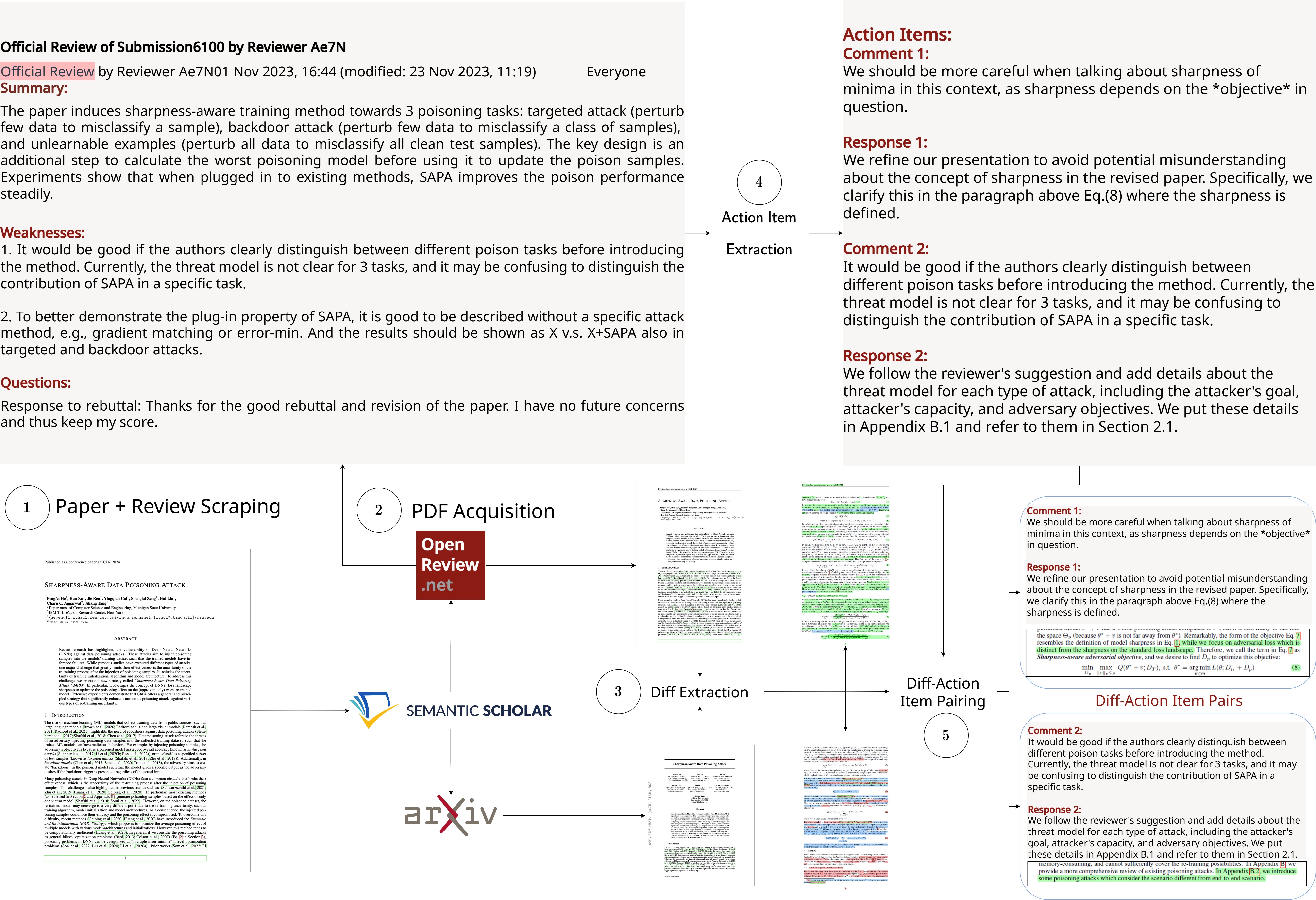}
    \caption{Block Diagram showing the various steps to produce the \datasetname{} dataset. (1) OpenReview is scraped to obtain submissions. (2) The submissions are then provided to Semantic Scholar to obtain the corresponding arXiv pre-submission PDFs. (3) Diffs are computed between the two versions. (4) Simultaneously the action items are extracted from the reviews, and they are (5) paired with the action items to produce the final version of \datasetname{}.}
    \label{fig:pipeline}
\end{figure}

\datasetname{} is collected by leveraging the review-rebuttal process on OpenReview. The back-and-forth dialogue between reviewers and authors produces a rich record of what changes were made to the paper and a rationale behind them. \datasetname{} collection proceeds in six stages: (1)~paper and review scraping, (2)~PDF acquisition, (3)~PDF Difference Computation, (4)~Action item extraction, (5)~human-annotation and filtering, (6)~Inter annotator agreement and dataset assembly.  Each stage is described in detail below and is exemplified by Figure~\ref{fig:pipeline}.

\subsubsection{Stage 1: Paper and Review Scraping}
The OpenReview API\footnote{\url{https://docs.openreview.net}} is used to obtain all submissions to the target venue, ICLR 2024, in compliance with the OpenReview terms of service \footnote{ \url{https://openreview.net/legal/terms}}. While \datasetname{} focuses specifically on ICLR, the process can be extended to any venue that publishes reviews on OpenReview. \datasetname{} is limited to work where reviewer comments are addressed in a camera-ready version of the paper. As a result, only the accepted papers are retained from the set of all submissions since the existence of a camera-ready version is guaranteed. For each accepted paper, the full reviewer-author dialogue is extracted by traversing the reply tree rooted at each \texttt{Official\_Review} note, recursively collecting every \texttt{Official\_Comment} reply and sorting the resulting thread chronologically. Then, each dialogue thread is stored as a structured record containing the initial review, the sequence of author and reviewer follow-up comments, and the anonymous reviewer identifier.

\subsubsection{Stage 2: PDF Acquisition}
The next stage is to obtain the PDFs before and after the review-rebuttal process. Since OpenReview does not make available the version of the paper that was submitted, only the final camera-ready version, the existence of each paper is first checked on arXiv\footnote{\url{https://arxiv.org}}. 

However, OpenReview and arXiv do not provide cross-referencing and operate as independent systems. Therefore, Semantic Scholar \cite{fricke2018semantic, kinney2025semanticscholaropendata} is used to link OpenReview papers to their arXiv preprints. Semantic Scholar indexes papers by title and exposes both the accepted venue and the arXiv ID (if it exists). 

The process starts by first obtaining the camera-ready paper from OpenReview. The PDF and the submission timestamp creation date (\texttt{cdate}) are recorded. Then, Semantic Scholar is queried with the paper title. A match is verified with the normalized Levenshtein distance between the returned title and the title used in the query. If the top result has a distance more than 5\%, the search result is rejected and the paper is discarded. The paper's arXiv identifier is resolved from the returned result from Semantic Scholar. Using the arXiv API, all versions of the paper are obtained. Each version's arXiv upload date is compared to the OpenReview \texttt{cdate} and only the most recent upload on or before the submission date is retained, ensuring that the downloaded version is the one closest to the version under review, excluding post-submission updates. Papers for which a version history cannot be obtained are excluded as a safeguard. All downloads and queries are performed with exponential-backoff retry logic to comply with rate limits.

\begin{lstlisting}[
  float=tbp,
  caption={Abbreviated diff-classification instance.
  Ellipses indicate omitted candidates. The diff at index 441 represents the revision to the manuscript made by the paper's authors in response to the reviewer comment.},
  label={lst:diff-schema},
  basicstyle=\ttfamily\scriptsize,
  breaklines=true,
  columns=fullflexible,
  frame=single,
  numbers=right,
  numberstyle=\tiny\color{gray},
  stepnumber=1,
  numbersep=6pt,
  xleftmargin=2pt,
  xrightmargin=2pt
]
{
  "paper_id": "xBfQZWeDRH",
  "action_item": {
    "comment": "The measure of training acceleration is based on the number of epochs...",
    "response": "We revised the manuscript to explain this comparison more explicitly."
  },
  "all_diffs": [
  {
    "diff_index": 0,
    "tag": "replace",
    "old": "Original Title Here",
    "new": "PUBLISHED AT ICLR 2024 ORIGINAL TITLE"
  },
  {
    "...": "diffs 1--440 omitted"
  },
  {
    "diff_index": 441,
    "tag": "replace",
    "old": "in the training procedure,",
    "new": "(64 vs. 256 epochs),",
    "page": 6
  },
  {
    "...": "diffs 442--1066 omitted"
  }
],
  "labels": [
    false, "...", true, "...", false
  ],
  "relevant_diff_indices": [441]
}
\end{lstlisting}

\subsubsection{PDF Difference Computation}

The next step is to compute the difference between the PDFs submitted to the venue and the post-submission version. To obtain the difference, a structured diff is computed between the two PDFs following the PDF-Diff repository \cite{ssibb_pdf_diff_viewer}. A \texttt{diff} is a contiguous text-level change representing the difference between two versions of a document, identifying content that was added, removed, replaced, or moved \cite{opengroup_diff}. First, both PDFs are parsed using PyMuPDF \cite{pymupdf}, followed by Git's histogram diff algorithm \footnote{https://git-scm.com/docs/diff-options.html\#Documentation/diff-options.txt - - -histogram} to obtain differences. The diff extractor operates at the text-block level, producing a list of edits, each tagged as \texttt{insert}, \texttt{delete}, or \texttt{replace}, together with the original and revised text spans, their surrounding context, page numbers, and word-level bounding boxes. The result is a list of differences between the original and modified versions of the paper, an example is available in Listing \ref{lst:diff-schema}.

\subsubsection{Action Item Extraction}
Reviewer-author dialogues often discuss multiple aspects of a paper. In addition to pointing out areas for improvement, reviewers typically ask clarifying questions regarding content. However, a significant part of the meta-reviewer's workflow is to distill this entire conversation into discrete \emph{action items}, i.e., pairs of a reviewer concern and the corresponding author commitment to correcting the specific change. 

To broadly filter the reviews into action items, \texttt{Gemma-3-27B-IT} (temperature $0.1$) \cite{team2025gemma} is provided with the review weaknesses, questions, and discussion thread and prompted to list the action items.  The prompt instructs the model to identify statements in which the authors explicitly commit to modifying the manuscript (e.g.\ ``We have revised Section~3\ldots'', ``We added an experiment\ldots'') and to extract each as a \texttt{(comment, response)} pair, prioritizing items that reference specific manuscript locations (section, table, equation or figure numbers). The full prompt is available in Listing \ref{app:action-item-prompt} in Appendix \ref{app:prompt-gemma}.

\subsubsection{Human Annotation and Filtering}

\paragraph{Comment Filtering}
Each LLM-extracted action item is presented to an annotator who labels it as \emph{keep}, \emph{maybe}, or \emph{discard}.  The filtering UI displays the reviewer comment and author response side by side, supports keyboard shortcuts for rapid annotation, and persists progress to a local cache file so that sessions can be interrupted and resumed.

\begin{figure}[t]
    \centering
    \includegraphics[width=0.9\linewidth]{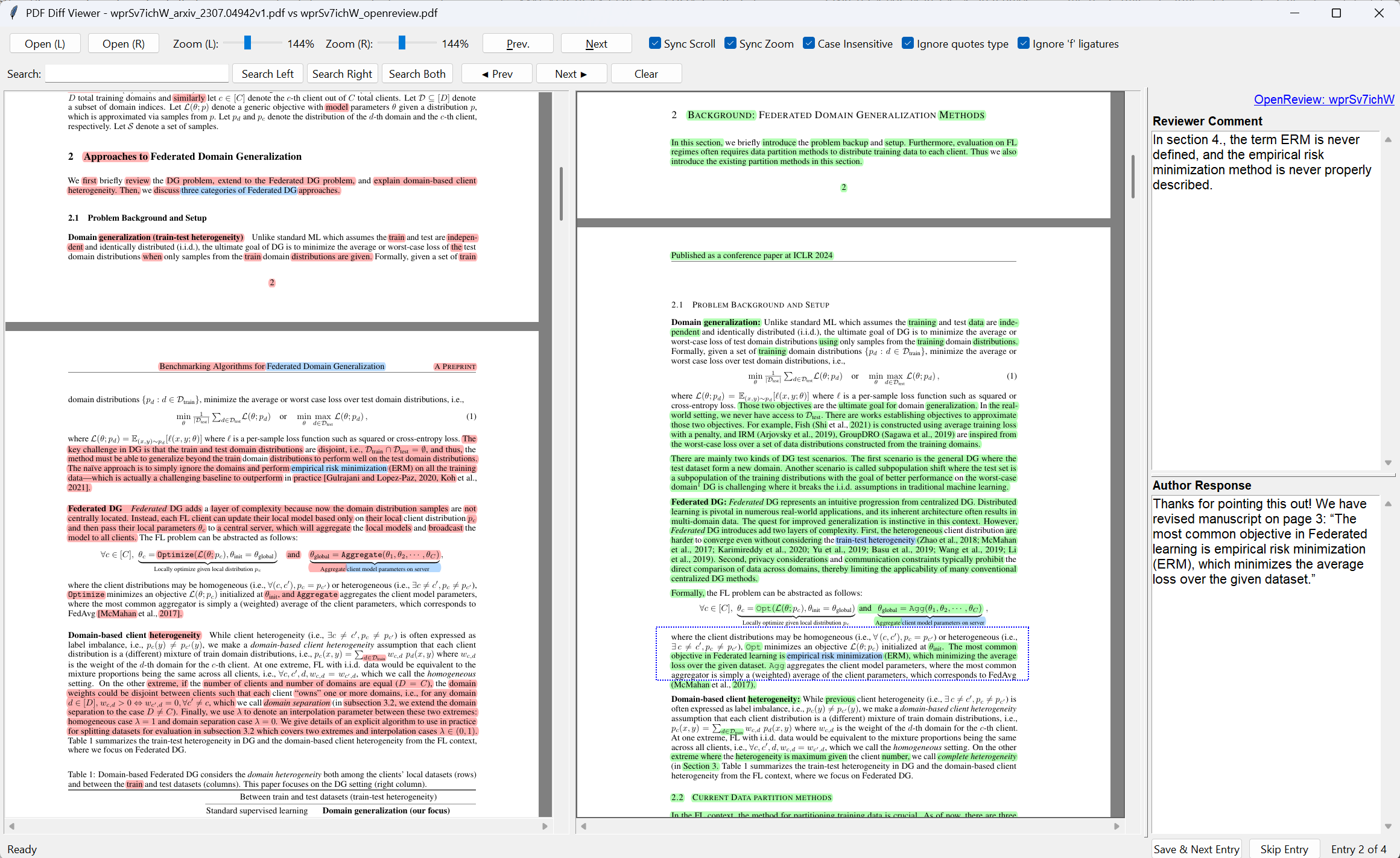}
    \caption{The annotation interface used to link diffs with reviewer comments and author responses.}
    \label{fig:Annotation Interface}
\end{figure}

\paragraph{Diff Linking}
Retained action items are routed to an interactive, side-by-side PDF viewer that renders the pre-revision (left pane) and post-revision (right pane) PDFs with colour-coded diff overlays.  Annotators read the reviewer concern and author response, then click on the PDF diff regions that correspond to the
described change.  Each click records the diff pane (\texttt{left}/\texttt{right}), page number, change type
(\texttt{insertion}/\texttt{deletion}/\texttt{moved}), the diff text, and its surrounding context. To amortize the cost of computing the diffs and rendering the PDF, the tool batches consecutive entries from the same paper. An example of the annotation window is available in Figure \ref{fig:Annotation Interface}. 

\label{sec:appendix-annotation}

\paragraph{Annotation framework}
A significant bottleneck for collecting paired datasets of edits and reviewer-author dialogue is a suitable interface to render and capture information accurately. To address this issue, we developed a Python-based annotation framework to identify reviewer-author action items and link them to the corresponding paper revisions. Annotators inspect the original and revised PDFs in a side-by-side diff-linking interface. The interface was implemented with Tkinter and PyMuPDF, with textual differences computed using Git's diff algorithm and a \texttt{difflib} fallback, following the pipeline outlined in \cite{ssibb_pdf_diff_viewer}. Insertions, deletions, replacements, and moved text are highlighted directly on the PDF pages, deletions appear in red, additions in green, moved text in blue. Annotators can navigate between changes, search either document, synchronize scrolling and zooming, and select relevant changes using Shift+Click or Shift+Drag. The interface displays the current reviewer concern and author response, groups action items by paper to reuse computed diffs, and provides controls to save, skip, or advance to the next item. Selected diffs, including their text, type, page, and surrounding context, are written to JSONL files, while automatic progress detection allows interrupted annotation sessions to resume from the last completed item. A link to the review on OpenReview is provided for additional context, should the annotator require it.  We will release the code to replicate the annotation framework upon acceptance. 

\subsubsection{Inter Annotator Agreement and Dataset Assembly}

Each action item is independently annotated by two annotators.  The linked \texttt{diffs} are then reconciled by taking the intersection of the sets of differences identified by each annotator. Only the \texttt{diffs} that appear in the intersection are retained. Two annotators are used to limit the time spent on annotation low while maximizing the number of cleaned data samples. Final statistics of the dataset are detailed in Table \ref{tab:dataset-statistics}. Annotators agreed on the exact set of diffs 37\% of the time and annotated sets had a mean Jaccard similarity of 50.1\%. Across the annotator pairs, binary agreement on whether an action item corresponded to at least one manuscript diff yielded a mean Cohen’s $\kappa$ of $0.369$, indicating fair agreement beyond chance. An example of a final datum in the dataset is provided in Listing \ref{lst:diff-schema}, and a description of the dataset fields is available in Appendix \ref{app:schema}. Additional details on the annotator agreement are provided in Appendix \ref{app:annotator}. 

\subsection{Diff Classification}
Diff classification is addressed using two approaches. The first is naive lexical retrieval using BM25 \cite{10.1145/2682862.2682863} and TF-IDF \cite{SprckJones2021ASI} to rank candidate diffs by their similarity to the action item (which serves as the query). The second is using neural approaches with language models. The language modeling-based approach utilizes a mix of open and closed source models. Using open-source models, we first experiment with an embedding-based approach that computes embeddings for every diff and action item followed by a classification head that classifies the diff as relevant or not. Next, we also explore a candidate-selection approach where the language model is tasked with selecting the diff(s) relevant to an action item from a list of candidates.  Because each paper contains approximately 1,000 candidate diffs on average,the diff pool is partitioned into contiguous, non-overlapping windows of up to 80 candidates. The number of candidates in the window was selected based on the validation split, and an analysis is provided in Appendix \ref{app:window} . Each candidate is represented by its global index, edit type, page number, original and revised text, and surrounding context. Language models then receive the action item and one candidate window and are tasked with returning the indices of all relevant diffs, or an empty list of none of the provided diffs are relevant. The prompt instructed the models to select only candidates in the provided window that directly implemented the action item. For each window, models return a JSON object of the form \texttt{\{"relevant\_diff\_indices": [3, 17]\}}, or an empty list when no candidate was relevant. Finally, the complete prediction for an action item is obtained by taking the union of the indices selected across all windows. 

Returned indices were validated against the candidates in the current window. Out-of-window indices were treated as hallucinations and discarded. The valid predictions from all windows belonging to an action item were then unioned to obtain its final predicted diff set. This set was compared with the human-annotated relevant indices to calculate precision, recall, and F1.

\subsection{Evaluation Metrics}

For each action item $i$, the predicted set of relevant diff
indices, $\hat{D}_i$, is compared with the human-annotated gold set, $D_i$. 
True positives, false positives, and false negatives are computed as
\[
\mathrm{TP}_i = |D_i \cap \hat{D}_i|,\qquad
\mathrm{FP}_i = |\hat{D}_i \setminus D_i|,\qquad
\mathrm{FN}_i = |D_i \setminus \hat{D}_i|.
\]

Given the true positives, false positives, and false negatives, Precision, Recall, and F1-score are computed at the micro and macro level. The micro-averaging pools decisions across all N action items while macro-averaging first computes each of the three metrics independently per sample then averages scores across items. 

\begin{table}[t]
\centering
\caption{Metag dataset statistics}
\label{tab:dataset-statistics}
\small
\begin{tabular}{lrrrrrr}
\toprule
Split & Samples & Papers & Diffs & Diffs/sample
      & Relevant/sample & Input tokens \\
\midrule
Train      & 248 & 93 & 219,569 & 885.4   & 6.28 & 2,880,154 \\
Validation & 41  & 22 & 41,445  & 1,010.9 & 6.66 & 2,971,307 \\
Test       & 60  & 22 & 62,772  & 1,046.2 & 8.23 & 4,471,357 \\
\midrule
Total      & 349 & 137 & 323,786 & 927.8 & 6.66 & 10,322,818 \\
\bottomrule
\end{tabular}
\end{table}

\begin{table*}[t]
\centering
\caption{Diff-classification results. LLM results report mean
$\pm$ sample standard deviation across three runs.}
\label{tab:diff-classification-results}
\setlength{\tabcolsep}{3.5pt}
\renewcommand{\arraystretch}{1.08}
\small
\begin{adjustbox}{max width=\textwidth}
\begin{tabular}{llccc:ccc}
\toprule
\rowcolor{gray!20}
& & \multicolumn{3}{c}{\textbf{Micro}} &
    \multicolumn{3}{c}{\textbf{Macro}} \\
\rowcolor{gray!20}
\textbf{Model} & \textbf{Setting} &
\textbf{P} & \textbf{R} & \textbf{F1} &
\textbf{P} & \textbf{R} & \textbf{F1} \\
\midrule

\rowcolor{gray!10}
\multicolumn{8}{l}{\textit{Validation}} \\

TF--IDF & top-$6$
& .098 & .088 & .092 & .098 & .290 & .123 \\

BM25 & top-$6$
& .114 & .103 & .108 & .114 & .354 & .146 \\

Gemma-3-27B-IT zero-shot & $w=50$
& .021 & .634 & .041 & .035 & \textbf{.652} & .058 \\

Gemma-3-27B-IT LoRA & $w=50$
& .116 & .297 & .167 & .026 & .064 & .029 \\

Gemma embedding + MLP & pairwise
& .108 & .088 & .097 & .105 & .187 & .087 \\

\addlinespace
GPT-5.6-Sol & $w=80$
& $.212 \pm .004$ & $\mathbf{.675 \pm .015}$
& $.322 \pm .006$
& $.330 \pm .020$ & $.609 \pm .030$
& $\mathbf{.363 \pm .024}$ \\

DeepSeek-V4-Pro & $w=80$
& $\mathbf{.381 \pm .101}$ & $.448 \pm .150$
& $\mathbf{.410 \pm .121}$
& $\mathbf{.373 \pm .023}$ & $.432 \pm .038$
& $.345 \pm .035$ \\

Kimi-K2.5 & $w=80$
& $.161 \pm .012$ & $.590 \pm .091$
& $.252 \pm .020$
& $.272 \pm .012$ & $.643 \pm .012$
& $.336 \pm .001$ \\

\midrule
\rowcolor{gray!10}
\multicolumn{8}{l}{\textit{Test}} \\

TF--IDF & top-$6$
& .114 & .083 & .096 & .114 & .226 & .127 \\

BM25 & top-$6$
& .117 & .085 & .098 & .117 & .263 & .136 \\

Gemma-3-27B-IT zero-shot & $w=50$
& .019 & \textbf{.601} & .037 & .026 & \textbf{.643} & .046 \\

Gemma-3-27B-IT LoRA & $w=50$
& .052 & .198 & .083 & .034 & .162 & .050 \\

Gemma embedding + MLP & pairwise
& .042 & .207 & .069 & .063 & .250 & .071 \\

\addlinespace
GPT-5.6-Sol & $w=80$
& $\mathbf{.270 \pm .004}$ & $.540 \pm .008$
& $\mathbf{.360 \pm .005}$
& $\mathbf{.396 \pm .018}$ & $.629 \pm .020$
& $\mathbf{.398 \pm .018}$ \\

DeepSeek-V4-Pro & $w=80$
& $.236 \pm .025$ & $.296 \pm .031$
& $.261 \pm .021$
& $.295 \pm .005$ & $.385 \pm .021$
& $.264 \pm .014$ \\

Kimi-K2.5 & $w=80$
& $.162 \pm .006$ & $.580 \pm .033$
& $.254 \pm .009$
& $.298 \pm .008$ & $.607 \pm .017$
& $.319 \pm .007$ \\

\bottomrule
\end{tabular}
\end{adjustbox}
\end{table*}

\section{Results}

Table~\ref{tab:diff-classification-results} reports the performance of the lexical and LLM-based approaches.

\paragraph{Lexical retrieval baselines.}
We evaluate BM25~\cite{10.1145/2682862.2682863} and unigram TF-IDF cosine similarity~\cite{SprckJones2021ASI}. The reviewer concern and author response are concatenated to form the query, while each candidate diff is represented by its original text, revised text, and surrounding context. The retrieval depth $k$ is selected using validation micro-F1, a complete sweep is reported in Appendix~\ref{app:optimal-k}. Both methods
select $k=6$, which is close to the average number of relevant diffs per action item.

The lexical baselines perform poorly overall. BM25 achieves micro-F1 scores of .108 on validation and .098 on test, while TF--IDF obtains .092 and .096, respectively. Their low recall indicates that relevant revisions often cannot be identified through direct lexical overlap alone. In many cases, the action item describes the intent of a revision
rather than repeating the language introduced in the manuscript.

\paragraph{LLM-based diff classification.}
We first evaluate \texttt{Gemma-3-27B-IT}~\cite{team2025gemma} using zero-shot prompting and LoRA fine-tuning~\cite{hu2021lora}. We also evaluate a pairwise classifier that uses mean-pooled \texttt{Gemma-3-27B-IT} embeddings to determine whether each candidate
diff is relevant. We further evaluate the hosted models
\texttt{GPT-5.6-Sol}~\cite{openai2026gpt56sol},
\texttt{DeepSeek-V4-Pro}~\cite{xu2026deepseek}, and
\texttt{Kimi-K2.5}~\cite{team2026kimi} using zero-shot prompting. The prompt is presented in Appendix \ref{app:llm-prompt}.

Because each action item is associated with approximately 1,000 candidate diffs, we partition the candidate pool into contiguous, non-overlapping windows. The \texttt{Gemma} experiments use 50 diffs per prompt, corresponding to a median estimated prompt length of approximately
3.1K tokens. For the hosted models, the window size is selected by evaluating \texttt{DeepSeek-V4-Pro} on validation windows ranging from 10 to 320 candidates. A window size of 80 obtains the highest mean validation micro-F1, while reducing the number of repeated instructions
and API calls. This setting produces median prompts of approximately 4.5K tokens and the complete window-size analysis is provided in Appendix~\ref{app:window}.

All hosted models receive identical prompts and candidate windows. \texttt{GPT-5.6-Sol} is accessed through the Responses API, whereas \texttt{DeepSeek} and \texttt{Kimi} use the Chat Completions API. Temperature, nucleus
sampling, reasoning effort, and other decoding parameters are left at their deployment defaults. Each model is evaluated in three independent runs, and we report the mean and sample standard deviation. Failed API or parsing requests are rerun and merged with the original predictions.

\paragraph{Main findings.}
\texttt{GPT-5.6-Sol} provides the strongest overall test performance, achieving a micro-F1 of $.360 \pm .005$ and a macro-F1 of $.398 \pm .018$. It combines relatively high precision ($.270 \pm .004$) with high recall ($.540 \pm .008$), yielding a better balance than the other hosted models.

On validation, \texttt{DeepSeek-V4-Pro} obtains the highest mean micro-F1 ($.410 \pm .121$), although its large standard deviation indicates substantial run-to-run variability. Its test micro-F1 decreases to $.261 \pm .021$. \texttt{Kimi-K2.5} achieves high recall on both validation and test, but its lower precision limits its test micro-F1 to $.254 \pm .009$.

The \texttt{Gemma-3-27B-IT} variants underperform the hosted models. Zero-shot \texttt{Gemma-3-27B-IT} achieves high recall (.634 on validation and .601 on test), but this behavior is driven by severe over-selection. It predicts over 200 diffs per action item despite only 6-8 being relevant on average. Consequently, its test micro-F1
is only .037. LoRA fine-tuning makes \texttt{Gemma-3-27B-IT} more selective and improves test micro-F1 to .083, while the
\texttt{Gemma-3-27B-IT} embedding-based MLP reaches .069. Although the LoRA-tuned \texttt{Gemma-3-27B-IT} slightly outperforms the lexical baselines on validation, \texttt{BM25} and \texttt{TF--IDF} perform
better on test. The overall difference in between performance on validation and test split is attributed to minor variations in distribution over the dataset splits. 

Overall, the results show that linking review action items to
manuscript revisions requires more than surface-level lexical
similarity. \texttt{GPT-5.6-Sol} provides the strongest and most stable test performance. 

\section{Conclusions and Future Work}
This paper presents \datasetname{}, a dataset to help build agentic capabilities to assist with meta-reviewing in the peer-reviewing process. \datasetname{} is collected by scraping reviews from OpenReview, identifying action items, and linking action items to specific changes made to scientific manuscripts as part of the review-rebuttal process. The resulting dataset consists of 349 high-quality human-reviewed samples, and is larger than prior datasets that collect paired edits linked with scientific reviews. \datasetname{} is benchmarked using lexical retrieval baselines and open and closed-source LLMs as well. \texttt{GPT-5.6-Sol} emerges as the most capable model on the task among the models evaluated, achieving an F1-score close to 0.40. 

\datasetname{} is collected from ICLR 2024 and the extension of the dataset to more venues is left to future work. Additional steps include further benchmarking with open-source models as context lengths of released models improve. \datasetname{} relies on linking camera-ready submissions on OpenReview with pre-prints on arXiv. Working closely with program organizers to make intermediate versions of under-review documents available can further improve research in dialogue-grounded iterative edits. 

\section{Limitations}
\label{sec:limitations}
\datasetname{} relies on accurately identifying the manuscript version submitted to the conference. However, papers may undergo multiple revisions throughout the review and rebuttal process, and these intermediate versions are
not always publicly available. Collaboration with conference organizers to release timestamped intermediate revisions would enable more fine-grained analysis of how manuscripts evolve in response to reviewer feedback and would support broader research on iterative scientific editing. A further limitation is that \datasetname{} is derived exclusively from ICLR 2024. Extending the dataset to additional venues, years, disciplines, and publication formats is therefore necessary to evaluate and improve its generalizability beyond a single machine-learning conference.

\bibliographystyle{plain}
\bibliography{refs}


\newpage
\appendix

\section{Prompt for Action Item Extraction}
\label{app:prompt-gemma}

The prompt for extracting action items as provided to \texttt{Gemma-3-27B-it} is provided in Listing \ref{lst:diff-schema}. 

\begin{lstlisting}[
  caption={Abbreviated diff-classification instance.
  Ellipses indicate omitted candidates.},
  label={app:action-item-prompt},
  basicstyle=\ttfamily\tiny,
  breaklines=true,
  columns=fullflexible,
  frame=single,
  numbers=right,
  numberstyle=\tiny\color{gray},
  stepnumber=1,
  numbersep=6pt,
  xleftmargin=2pt,
  xrightmargin=2pt,
]
    You are an expert at analyzing scientific paper reviews. 
## Task: 
You will be given:
1. **Review**: A review of a scientific paper submitted to a top tier conference in machine learning
2. **Dialogue**: Dialogue between the authors of the paper and the reviewers. 
## Instructions: 
Analyze the review and the dialogue to identify action items for authors to fix in their paper. 
Look for statements like typos and grammatical errors that reviewers point out and ask to be fixed, and statements from the authors stating that they will fix content in the paper. Examples of this include "We have updated the manuscript to clarify this point.", "We have added details in the revised version", look for any words indicating that the authors have fixed the reviewer concerns. Your responses should **only** include those cases with phrases like "updated the manuscript", "revised the description", or "edited the paper" indicating that the change was made. Prioritize those responses where there is a reference to a specific section in the paper such as "Section 3.1", "Equation 4", etc which can help point to where those changes were made. The text should be directly extracted from the dialogues, do not write in third person. For all examples of this, make sure to **paraphrase** the reviewer comment and the author response, do not hallucinate content. Provide your output as a list of JSON entities in the following format: 
## Output: 
{
    "entities": [
        {
        "comment": <Comment from Review asking something to be fixed>,
        "response": <Response from Dialogue stating how the issue will be fixed>,
        },
        {
        "comment": <Another comment from Reviewe asking something to be fixed>,
        "response": <Another response from the Dialogue stating how the issue will be fixed>,
        },
    ]
}
## Example Input: 
## Review:
1. As this research utilized a named entity recognition model to extract keywords, it is possible that the NER model can extract privacy information such as patient names. Is there any filtering or postprocessing step to avoid that? In addition, it is not guaranteed that NER system will never extract sensitive patient information; for example, if the NER system incorrectly extracts a patient's address as a symptom, then the address may be leaked to LLM. Although it is very rare, it is still necessary to comment on this.
2. As the LLM already provides a preliminary decision, I am curious about the performance if we only feed the preliminary decision from LLM to SLM. It is worth knowing which part of the LLM-generated information improves the SLM most.
3. The related work section need to discuss more LLM application in the clinical area, especially the knowledge-enhanced LLM in clinical settings. For example, paper "Qualifying Chinese Medical Licensing Examination with Knowledge Enhanced Generative Pre-training Model." also utilized external knowledge for clinical questions.
4. By adding the LLM-generated content, will the new concatenated input be too long and out of the word window in SLM? How do you deal with the long content problem? By adding the LLM-generated content, will the new concatenated input be too long and out of the word window in SLM? How do you deal with the long content problem?
## Dialogue:
We appreciate the insightful feedback and comments from the reviewer. Their positive observations about the novelty and thoroughness of our experiments are very encouraging. We have addressed your concerns in our response.\n\n**1. Concerns on privacy preserving in practical usage.**\nThe data we utilized in experiments have already undergone post-processing; however, even well-processed data cannot be directly shared with third parties in a real-hospital setting. Here, we adopt NER methods directly, solely for automation, to show that LLM can be utilized as a medical database to query knowledge under privacy-restricted scenarios. Practically, we can leverage de-identification models and rules to remove personal information and then extract medical keywords to query third-party LLMs for auxiliary knowledge generation. In this paper, we take an initial step to discuss the significant privacy-preserving situations in the medical domain and demonstrate the promising results of utilizing LLM to improve SLM while mitigating privacy concerns.\n\n **2.Question about what SLM learns for decision making.**We feed preliminary decisions (PD) as context into SLM with backbone BioLinkBert-Base on three datasets. Three separate runs for each setting are conducted and the average results along with the standard deviation are reported. The results are shown in the Table below. |MEDQA | HeadQA        | MEDMCQA
 SLM w PD   | 47.21  0.31  | 53.64 1.09  | 45.42 0.17  |
| FTC        | 50.17  0.42  | 61.35 0.16  | 49.20 0.45  |FTC, which integrates extensive medical knowledge into the decision-making, shows a consistent and significant improvement over the SLM that only uses PD for context. These findings underscore the valuable contribution of leveraging comprehensive medical knowledge, provided by LLM, in enhancing the medical decision-making capabilities.**3. Suggestion about related work in LLM application in the clinical domain.**\n\nThanks for the suggestion in the related work. We will add the suggested work into the related work section in the revision.**4. Question abut address long medical context generated by LLM.**We utilize the Fusion-in-Decoder [1] approach in our general domain experiments. This strategy is also effective for encoding long contexts. It works by dividing the input into smaller passages, encoding each one separately, and then combining the encoded representations for decision-making.[1] Izacard et al. ( 2020) Leveraging passage retrieval with generative models for open domain question answering\nThanks for replying. Based on the response, I would like to keep my original score.

## Output: 
{
    "entities": [
        {
        "comment": "The related work section need to discuss more LLM application in the clinical area, especially the knowledge-enhanced LLM in clinical settings. For example, paper "Qualifying Chinese Medical Licensing Examination with Knowledge Enhanced Generative Pre-training Model." also utilized external knowledge for clinical questions.",
        "response": "Thanks for the suggestion in the related work. We will add the suggested work into the related work section in the revision.",
        },
    ]
}

\end{lstlisting}

\section{Dataset Schema}
\label{app:schema}

Each entry corresponds to one (action item, paper) pair:


\begin{table}[t]
\centering
\caption{Schema of a diff-classification instance.}
\label{tab:diff-schema}
\small
\begin{tabularx}{\columnwidth}{@{}l l X@{}}
\toprule
Field & Type & Description \\
\midrule
\texttt{paper\_id}
  & String
  & Unique paper identifier. \\

\texttt{action\_item}
  & Object
  & Reviewer concern and author's description of the revision. \\

\texttt{all\_diffs}
  & List
  & Candidate changes between the original and revised paper. \\

\texttt{diff\_index}
  & Integer
  & Candidate's global index within the paper. \\

\texttt{tag}
  & String
  & Edit operation, such as insertion, deletion, or replacement. \\

\texttt{text\_pdf1}
  & String
  & Text from the original paper. \\

\texttt{text\_pdf2}
  & String
  & Corresponding text from the revised paper. \\

\texttt{context\_*}
  & String
  & Text immediately before or after the edit. \\

\texttt{page\_nums\_*}
  & List
  & Page locations in the original or revised paper. \\

\texttt{labels}
  & Boolean list
  & Diff-level relevance labels aligned with \texttt{all\_diffs}. \\

\texttt{relevant\_diff\_indices}
  & Integer list
  & Global indices of the diffs implementing the action item. \\
\bottomrule
\end{tabularx}
\end{table}


The \texttt{labels} array is aligned with \texttt{all\_diffs}: \texttt{labels[i]} is \texttt{true} if \texttt{all\_diffs[i]} is relevant to the action item. The \texttt{relevant\_diff\_indices} field lists the indices of all true labels for convenience. The schema is provided in Table \ref{tab:diff-schema}.

\section{Annotator Details}
\label{app:annotator}
Eleven annotators assisted with the labeling effort and the  annotation effort was between 2 and 3 hours depending on annotator. Each data sample was annotated by 2 annotators, with a sample making it into the final dataset if both annotators agreed on the relevant diff. Table \ref{tab:inter-annotator-agreement} reports statistics on inter-annotator agreement. Annotators were compensated with co-authorship of the publication. In general, across 575 doubly annotated action items, annotators selected identical sets in $37$\% of cases. Partial agreement was higher, with a mean Jaccard overlap of $50.1$\%. Binary agreement on whether an action item corresponded to any manuscript diff yielded a mean pairwise Cohen's $\kappa$ of $0.369$. 

Of the 575 items, both annotators selected at least one diff for 404 items, neither selected a diff for 59, and only one selected a diff for 112. The variance in this metric indicates that there is subjectivity in the difference annotation in about $20$\% of cases. That $10.3$\% of cases did not have a relevant diff indicates that authors do not always commit to making changes though they are outlined during the review process, consistent with findings in prior work \cite{darcy-etal-2024-aries}.

\begin{table}[h]
\centering
\caption{Inter-annotator agreement over 575 doubly annotated
action items. Minimum and maximum values are computed across
annotator pairs.}
\label{tab:inter-annotator-agreement}
\begin{tabular}{lrrr}
\toprule
Metric & Overall & Minimum & Maximum \\
\midrule
Exact diff-set match & 37.0\% & 25.0\% & 46.9\% \\
Mean Jaccard          & 50.1\% & 41.0\% & 61.4\% \\
Pooled Dice           & 49.7\% & 44.8\% & 62.2\% \\
Cohen's $\kappa$      & 0.369  & 0.047  & 0.601 \\
\bottomrule
\end{tabular}
\end{table}

\begin{table}[h]
\centering
\caption{Agreement on whether an action item was linked to any diff,
computed over 575 doubly annotated items.}
\label{tab:diff-presence-agreement}
\begin{tabular}{lrr}
\toprule
Annotation outcome & Count & Percentage \\
\midrule
Neither annotator identified a diff & 59 & 10.3\% \\
Only one annotator identified a diff & 112 & 19.5\% \\
Both annotators identified a diff & 404 & 70.3\% \\
\bottomrule
\end{tabular}
\end{table}

\section{Windowing}
\label{app:window}
\begin{figure*}[t]
  \centering
  \includegraphics[width=0.7\textwidth]
    {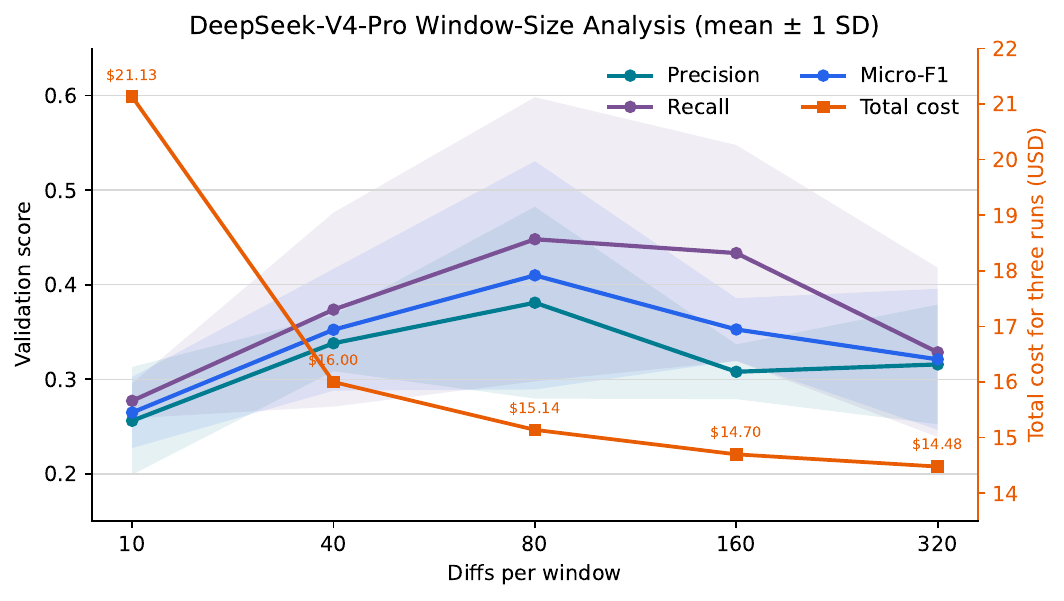}
  \caption{DeepSeek-V4-Pro validation performance and cost across
  window sizes. Shaded regions show $\pm1$ sample standard deviation
  across three runs.}
  \label{fig:window-size}
\end{figure*}

To select the optimal window size of the number of diffs to pass in a single prompt, \texttt{DeepSeek-V4-Pro} was swept over values $\{10, 40, 80, 160, 320\}$. As the window size increases, the number of prompts to evaluate and cost of evaluation decreases. On the other hand, the more diffs provided to the model, the more granularity available. Figure \ref{fig:window-size} reports the sweep over the parameters. A window size of $80$ resulted in optimal performance on the validation set, and costs roughly the same as the cheapest setting. 

\section{Hyperparamter Settings}
\label{app:optimal-k}

\begin{figure*}[t]
  \centering
  \includegraphics[width=\textwidth]
    {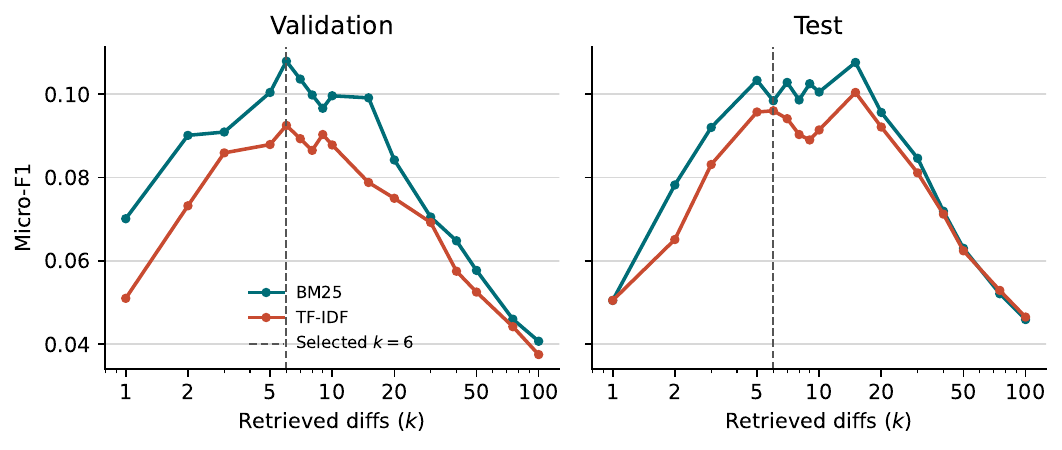}
  \caption{BM25 and TF--IDF micro-F1 across retrieval cutoffs.
  The dashed line marks the validation-selected $k=6$.}
  \label{fig:lexical-k-sweep}
\end{figure*}

\paragraph{Optimal $k$ for BM25 and TF-IDF: }
The optimal value for $k$ to be used in the BM25 and TF-IDF experiments were obtained by sweeping values from 1 to 100 on the validation set. The best value was obtained at $k=6$, and can be observed in Figure \ref{fig:lexical-k-sweep}. 

\paragraph{Gemma hyperparameters: }
We evaluated \texttt{Gemma-3-27B-IT} using zero-shot prompting, LoRA fine-tuning, and an embedding-based classifier. For the generative experiments, candidate diffs were grouped into batches of 50 after removing trivial formatting changes. Decoding used temperature $0.1$, nucleus-sampling probability $p=0.95$, and a maximum of 1,024 generated tokens. Zero-shot inference used a batch size of 16.

For fine-tuning, LoRA adapters were trained for three epochs with learning rate $2\times10^{-5}$. LoRA rank $r=8$, $\alpha=16$, and dropout $0.05$. The per-device batch size was 1 with four gradient-accumulation steps, giving an effective batch size of 4. Input sequences were truncated to 4,096 tokens.

For the embedding experiment, Gemma was loaded using 4-bit NF4 quantization with double quantization and \texttt{bfloat16} computation. Action items and diffs were independently truncated to 512 tokens, and their 5,376-dimensional representations were obtained by mean-pooling the final hidden layer over non-padding tokens. A three-layer MLP received the concatenation of the action-item embedding, diff embedding, and their element-wise product. The classifier used a hidden dimension of 256, dropout $0.3$, 10:1 negative sampling, and a positive-class weight of 10. Eight Nvidia A40 GPUs with 48GB GPU memory were used for LLM Experiments. 

Azure OpenAI was used for API access to closed-source LLMs, and default settings for reasoning effort, random seed, and output length were used. The total experimental cost of running the experiments detailed in this work using closed source LLMs totals USD 534.57.

\section{LLM Prompt}
\label{app:llm-prompt}
The prompt used in the LLM prompting approaches is provided in Listing \ref{lst:hosted-llm-prompt}. 

\begin{lstlisting}[
  float=tbp,
  caption={Zero-shot prompt shared by
  \texttt{GPT-5.6-Sol}, \texttt{DeepSeek-V4-Pro}, and
  \texttt{Kimi-K2.5}. Bracketed fields denote values populated
  for each action item and candidate window.},
  label={lst:hosted-llm-prompt},
  basicstyle=\ttfamily\scriptsize,
  breaklines=true,
  breakatwhitespace=false,
  columns=fullflexible,
  frame=single,
  numbers=right,
  numberstyle=\tiny\color{gray},
  stepnumber=1,
  numbersep=6pt,
  xleftmargin=2pt,
  xrightmargin=12pt
]
You are an expert at analyzing scientific paper revisions.

## Task
You are given one action item from a reviewer-author
discussion (a reviewer concern plus the author's response
describing a change they made) and a batch of PDF diffs
between the original and revised versions of the paper.

Identify which diffs implement the change described in the
action item.

## Guidelines
- A diff is relevant only if its content directly implements
  the described change.
- Look for matching keywords, section/table/figure references,
  or specific text named in the response.
- Most diffs are NOT relevant. Do not guess.
- Formatting-only diffs (citation style, capitalization, line
  breaks) are not relevant unless the action item is
  specifically about formatting.

## Output Format
Return a JSON object with a single key
"relevant_diff_indices" holding a list of diff_index values,
e.g. {"relevant_diff_indices": [3, 17, 42]}
Return {"relevant_diff_indices": []} if no diff in this batch
is relevant.

## Action Item
Reviewer concern: [REVIEWER CONCERN]
Author response: [AUTHOR RESPONSE]

## Diffs
[diff_index=INDEX] tag=TAG moved=MOVED page=PAGE
  context_before: [PRECEDING CONTEXT]
  old: [ORIGINAL TEXT]
  new: [REVISED TEXT]
  context_after: [FOLLOWING CONTEXT]

[REPEATED FOR EACH DIFF IN THE WINDOW]

## Output
\end{lstlisting}

\section{LLM Usage}
\label{app:llm-usage}
Language models were used to proofread the manuscript and assist with improving the overall writing. Coding assistants were used to assist with code development. 


\newpage
\section*{NeurIPS Paper Checklist}

\begin{enumerate}

\item {\bf Claims}
    \item[] Question: Do the main claims made in the abstract and introduction accurately reflect the paper's contributions and scope?
    \item[] Answer: \answerYes{} 
    \item[] Justification: The abstract and the introduction state the paper's contributions and scope. See Section \ref{sec:introduction}

\item {\bf Limitations}
    \item[] Question: Does the paper discuss the limitations of the work performed by the authors?
    \item[] Answer: \answerYes{} 
    \item[] Justification: Provided in Section \ref{sec:limitations}. 

\item {\bf Theory assumptions and proofs}
    \item[] Question: For each theoretical result, does the paper provide the full set of assumptions and a complete (and correct) proof?
    \item[] Answer: \answerNA{} 
    \item[] Justification: No theoretical proofs are presented in this work.

    \item {\bf Experimental result reproducibility}
    \item[] Question: Does the paper fully disclose all the information needed to reproduce the main experimental results of the paper to the extent that it affects the main claims and/or conclusions of the paper (regardless of whether the code and data are provided or not)?
    \item[] Answer: \answerYes{} 
    \item[] Justification: Hyperparameters and prompts are provided in the paper. 

\item {\bf Open access to data and code}
    \item[] Question: Does the paper provide open access to the data and code, with sufficient instructions to faithfully reproduce the main experimental results, as described in supplemental material?
    \item[] Answer: \answerNo{} 
    \item[] Justification: The code and models will be released post acceptance. Dataset examples are provided in the paper and further examples can be made available during the review process but the final data and code will be released post acceptance. 

\item {\bf Experimental setting/details}
    \item[] Question: Does the paper specify all the training and test details (e.g., data splits, hyperparameters, how they were chosen, type of optimizer) necessary to understand the results?
    \item[] Answer: \answerYes{} 
    \item[] Justification: See Appendix \ref{app:prompt-gemma},  \ref{app:window}, \ref{app:optimal-k}, . 

\item {\bf Experiment statistical significance}
    \item[] Question: Does the paper report error bars suitably and correctly defined or other appropriate information about the statistical significance of the experiments?
    \item[] Answer: \answerYes{} 
    \item[] Justification: Table \ref{tab:diff-classification-results} reports standard deviation, Figure \ref{fig:window-size} reports error bars. 

\item {\bf Experiments compute resources}
    \item[] Question: For each experiment, does the paper provide sufficient information on the computer resources (type of compute workers, memory, time of execution) needed to reproduce the experiments?
    \item[] Answer: \answerYes{} 
    \item[] Justification: See Appendix \ref{app:optimal-k}. 
    
\item {\bf Code of ethics}
    \item[] Question: Does the research conducted in the paper conform, in every respect, with the NeurIPS Code of Ethics \url{https://neurips.cc/public/EthicsGuidelines}?
    \item[] Answer: \answerYes{} 
    \item[] Justification: We have reviewed the code. 

\item {\bf Broader impacts}
    \item[] Question: Does the paper discuss both potential positive societal impacts and negative societal impacts of the work performed?
    \item[] Answer: \answerYes{} 
    \item[] Justification: Section \ref{sec:introduction} introduces the impact of the released dataset for the scientific community. 
    
\item {\bf Safeguards}
    \item[] Question: Does the paper describe safeguards that have been put in place for responsible release of data or models that have a high risk for misuse (e.g., pre-trained language models, image generators, or scraped datasets)?
    \item[] Answer: \answerNA{} 
    \item[] Justification: 

\item {\bf Licenses for existing assets}
    \item[] Question: Are the creators or original owners of assets (e.g., code, data, models), used in the paper, properly credited and are the license and terms of use explicitly mentioned and properly respected?
    \item[] Answer: \answerYes{}{} 
    \item[] Justification: Section \ref{sec:introduction} outlines the license as CDLA-2.0. 

\item {\bf New assets}
    \item[] Question: Are new assets introduced in the paper well documented and is the documentation provided alongside the assets?
    \item[] Answer: \answerYes{} 
    \item[] Justification: The dataset will be released post acceptance. Examples of the dataset are provided in Figure \ref{fig:teaser} and in Listing \ref{lst:diff-schema}. 

\item {\bf Crowdsourcing and research with human subjects}
    \item[] Question: For crowdsourcing experiments and research with human subjects, does the paper include the full text of instructions given to participants and screenshots, if applicable, as well as details about compensation (if any)? 
    \item[] Answer: \answerYes{} 
    \item[] Justification: The annotation framework is provided in Figure \ref{fig:Annotation Interface} and annotator information is provided in Section \ref{sec:appendix-annotation} and Appendix \ref{app:action-item-prompt}. 

\item {\bf Institutional review board (IRB) approvals or equivalent for research with human subjects}
    \item[] Question: Does the paper describe potential risks incurred by study participants, whether such risks were disclosed to the subjects, and whether Institutional Review Board (IRB) approvals (or an equivalent approval/review based on the requirements of your country or institution) were obtained?
    \item[] Answer: \answerNA{} 
    \item[] Justification: 

\item {\bf Declaration of LLM usage}
    \item[] Question: Does the paper describe the usage of LLMs if it is an important, original, or non-standard component of the core methods in this research? Note that if the LLM is used only for writing, editing, or formatting purposes and does \emph{not} impact the core methodology, scientific rigor, or originality of the research, declaration is not required.
    \item[] Answer: \answerYes{} 
    \item[] Justification: See section \ref{app:llm-usage}

\end{enumerate}

\end{document}